\documentclass[conference]{IEEEtran}
\IEEEoverridecommandlockouts
\usepackage{cite}
\usepackage{amsmath,amssymb,amsfonts}
\usepackage{algorithmic}
\usepackage{graphicx}
\usepackage{textcomp}
\usepackage{xcolor}
\def\BibTeX{{\rm B\kern-.05em{\sc i\kern-.025em b}\kern-.08em
    T\kern-.1667em\lower.7ex\hbox{E}\kern-.125emX}}
\begin{document}

\title{The Brain-Inspired Cooperative Shared Control Framework for Brain-Machine Interface\\

}

\author{

\IEEEauthorblockN{1\textsuperscript{st} Junjie Yang}
\IEEEauthorblockA{\textit{Shenzhen Institute of Advanced Technology, CAS} \\
\textit{University of Chinese Academy of Sciences}\\
Shenzhen, China \\
jj.yang@siat.ac.cn}
\and
\IEEEauthorblockN{2\textsuperscript{nd} Ling Liu}
\IEEEauthorblockA{\textit{Shenzhen Institute of Advanced Technology, CAS} \\
\textit{University of Chinese Academy of Sciences}\\
Shenzhen, China \\
ling.liu1@siat.ac.cn}
\and
\IEEEauthorblockN{3\textsuperscript{rd} Shengjie Zheng}
\IEEEauthorblockA{\textit{Shenzhen Institute of Advanced Technology, CAS} \\
\textit{University of Chinese Academy of Sciences}\\
Shenzhen, China \\
sj.zheng@siat.ac.cn}
\and
\IEEEauthorblockN{4\textsuperscript{th} Lang Qian}
\IEEEauthorblockA{\textit{Tsinghua Shenzhen International Graduate School} \\
\textit{Tsinghua University}\\
Shenzhen, China \\
ql20@mails.tsinghua.edu.cn}
\and
\IEEEauthorblockN{5\textsuperscript{th} Gang Gao}
\IEEEauthorblockA{\textit{Shenzhen Institute of Advanced Technology, CAS} \\
\textit{University of Chinese Academy of Sciences}\\
Shenzhen, China \\
gang.gao@siat.ac.cn}
\and
\IEEEauthorblockN{6\textsuperscript{th} Xin Chen}
\IEEEauthorblockA{\textit{Shenzhen Institute of Advanced Technology, CAS} \\
\textit{University of Chinese Academy of Sciences}\\
Shenzhen, China \\
xjy107033@siat.ac.cn}
\and
\IEEEauthorblockN{7\textsuperscript{th} Xiaojian Li}
\IEEEauthorblockA{\textit{Shenzhen Institute of Advanced Technology, CAS} \\
\textit{University of Chinese Academy of Sciences}\\
Shenzhen, China \\
xj.li@siat.ac.cn}
}

\maketitle

\begin{abstract}
In brain-machine interface (BMI) applications, a key challenge is the low information content and high noise level in neural signals, severely affecting stable robotic control. To address this challenge, we proposes a cooperative shared control framework based on brain-inspired intelligence, where control signals are decoded from neural activity, and the robot handles the fine control. This allows for a combination of flexible and adaptive interaction control between the robot and the brain, making intricate human-robot collaboration feasible. The proposed framework utilizes spiking neural networks (SNNs) for controlling robotic arm and wheel, including speed and steering. While full integration of the system remains a future goal, individual modules for robotic arm control, object tracking, and map generation have been successfully implemented. The framework is expected to significantly enhance the performance of BMI. In practical settings, the BMI with cooperative shared control, utilizing a brain-inspired algorithm, will greatly enhance the potential for clinical applications.

\end{abstract}


\section{Introduction}
Human-computer interaction (HCI) has advanced significantly, providing interfaces for communication between humans and robots, particularly in service robots, to improve daily convenience \cite{collinger2013high,hochberg2012reach,nuyujukian2018cortical,willett2021high,rajeswaran2021neural,simeral2011neural,mcfarland2008emulation}. However, its effectiveness is limited due to low information density, especially in complex, unstructured environments \cite{pfeifer2007self}. 

Brain-machine interfaces (BMI) represent a more advanced form of HCI, capturing neural signals to control robots and other systems, offering great potential for applications such as aiding paralyzed patients \cite{DawnMTaylor2002DirectCC,taylor2002direct,donoghue2002brain,chapin1999real,wessberg2000real,carmena2003learning}. Despite the larger information capture potential through additional electrodes, BMIs face challenges like low signal quality, high noise, and instability, which limits continuous control \cite{chaudhary2016brain,johansson2009coding}. This highlights the need for intelligent robots capable of interpreting complex instructions in collaboration with BMI.

The proposed framework differentiates itself from prior shared control techniques by introducing a conceptual collaborative control approach that integrates brain-inspired intelligence. The key idea is to use shared control strategies leveraging spiking neural networks (SNNs) to enhance the fluency of control, thus bridging the gap between human intentions and robotic execution. Unlike traditional shared control methods that often rely solely on pre-programmed responses or fixed control loops, this study introduces a flexible, cooperative control strategy that adapts dynamically to user intentions, providing a more fluid interaction.

Closed-loop BMI systems enable robots to interact with the brain to learn human preferences, supporting decision-making and task execution. Such systems leverage human-like reasoning and can fuse abstract brain commands with concrete control inputs, enhancing task performance and enabling humans to harness more of their cognitive potential.

Collaborative control algorithms are essential for BMIs in clinical settings since behavior involves not only the sensorimotor cortex but also subcortical regions like the cerebellum and brainstem \cite{vogel2015assistive,katyal2014collaborative}. These regions control instinctive actions, such as maintaining balance, which bypasses conscious decision-making \cite{yonekura2020spike}. In this study, the sensorimotor cortex is responsible for decoding precise motor commands, while subcortical regions like the cerebellum and brainstem regulate reflexive actions, ensuring the cooperative shared control system remains adaptive and responsive to environmental changes.

In this paper, we propose a conceptual collaborative shared control strategy for brain-controlled robots. The work aims to lay a foundation for future brain-machine interface solutions by providing a novel control concept and testing its components in isolation to demonstrate feasibility. The results demonstrate the potential for BMIs to enhance robot precision and control in complex environments, paving the way for intelligent robot systems capable of human-like cognitive abilities.

\section{Methods}
This section outlines the shared control strategies for robotic tasks assisting the brain-machine interface (BMI), including robotic arm grasping and wheel control. First, we describe the computer vision algorithm for object detection, followed by the robotic arm grasping system, and finally the wheel speed and orientation control algorithms.

\subsection{System Specification}

\begin{figure}[ht]
\centering
\includegraphics[width=\linewidth]{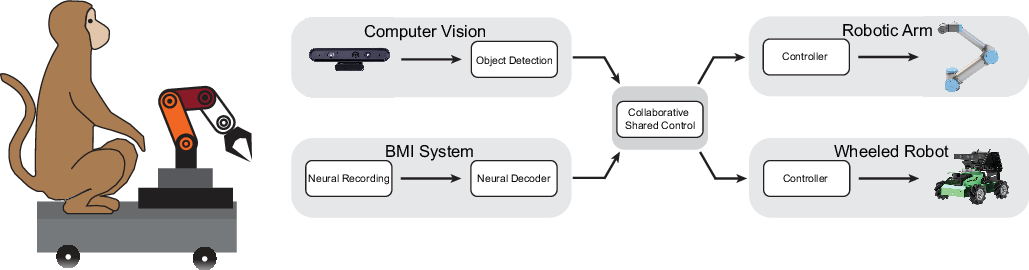}
\caption{Brain-machine interface cooperative shared control system}
\label{fig:bci_system}
\end{figure}

The proposed system consists of a computer vision subsystem, a robotic arm controlled by spiking neural networks, and mobile wheeled robots also driven by SNN. The computer vision module handles object detection, while the SNN-based algorithm controls the robotic arm, moving it from the initial position to the target, and moves objects to predefined areas. The SNN algorithm also enables the wheeled robots to navigate toward the target efficiently and generate a dense semantic map. Fig. \ref{fig:bci_system} shows the system schematic.

\subsection{Computer vision system}
We employ the YOLOv5 algorithm for object detection, balancing speed and accuracy. An RGB camera captures images from a fixed position above the robots and is calibrated using the Zhang method \cite{zhang2000flexible} to obtain camera parameters.

For the robotic arm, binocular cameras provide the target's image coordinates, which are processed using YOLOv5s. Newton’s method \cite{wedderburn1974quasi} approximates the object's real-world coordinates, allowing the arm to grasp the target.

For the wheeled robots, the Orbbec Astra Pro depth camera handles 3D reconstruction and SLAM. The system uses object detection to control the robot's movement based on the target’s position on the screen and depth information.

\begin{figure}[ht]
\centering
\includegraphics[width=\linewidth]{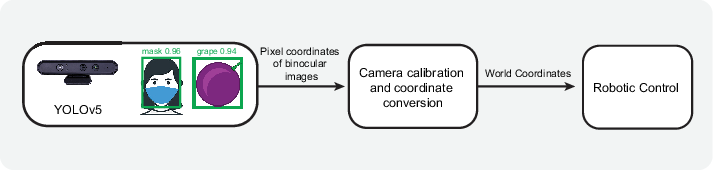}
\caption{The computer vision system framework.}
\label{fig:vision_system}
\end{figure}

We chose YOLOv5s for its efficient performance with minimal resources. The system's training process includes data augmentation and testing for optimal speed and accuracy on low-computation platforms. Fig. \ref{fig:vision_system} illustrates the system framework.

\subsection{Robotic Arm Grasping System}

\begin{figure}[ht]
\centering
\includegraphics[width=\linewidth]{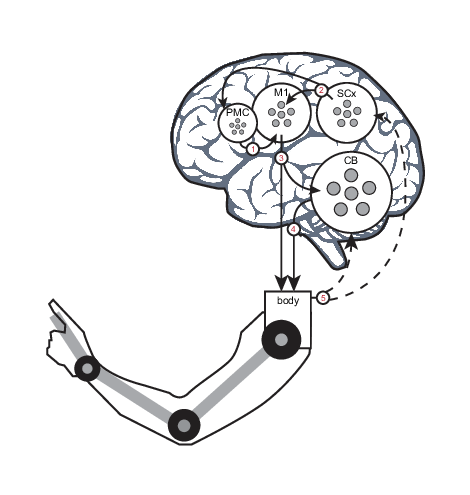}
\caption{An overview of the REACH model\cite{dewolf2016spiking}, shown controlling a three-link arm. Numbers identify major communication pathways. Dashed lines indicate closed-loop feedback signals generated from the senses. The image modified from Travis et al \cite{dewolf2016spiking}.}
\label{fig:reach}
\end{figure}

The robotic arm control is based on the recurrent error-driven adaptive control hierarchy (REACH) model proposed by Travis et al. \cite{dewolf2016spiking}. This model, which utilizes spiking neural networks, closely mimics biological motor control for robotic arm movement, as shown in Fig. \ref{fig:reach}.

The model integrates joint coordinates, velocities, and forces such as inertia, Coriolis effects, and gravitational forces. An adaptive correction signal is generated using SNN to adjust the control torque applied to each joint, allowing precise trajectory following.

The spiking neural network is built using the Neural Network Engineering Framework (NEF) and employs the Prescribed Error Sensitivity (PES) learning rule to adjust the control signals in real-time. The network is composed of 1000 Leaky Integrate and Fire (LIF) neurons, which process input data and generate motor control commands. The system continuously minimizes error signals through online learning, adapting the control strategy as the task evolves.

\begin{figure}[ht]
\centering
\includegraphics[width=\linewidth]{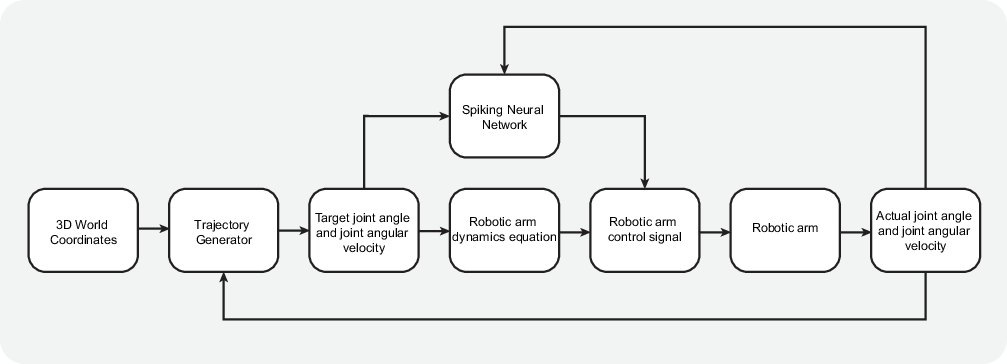}
\caption{The robotic arm control flow}
\label{fig:Robot_arm_control}
\end{figure}

This approach ensures efficient and biologically inspired control of the robotic arm, allowing it to perform precise grasping tasks. The entire control framework is shown in Fig. \ref{fig:Robot_arm_control}.

\subsection{Wheel Speed Control}

Mobile robots often use differential drive, which can result in instability and difficulties maintaining a straight trajectory due to factors such as motor inconsistencies, wheel slippage, and obstacles. Open-loop control is insufficient for correcting these deviations, necessitating closed-loop control to provide feedback for real-time adjustments.

\begin{figure}[ht]
\centering
\includegraphics[width=\linewidth]{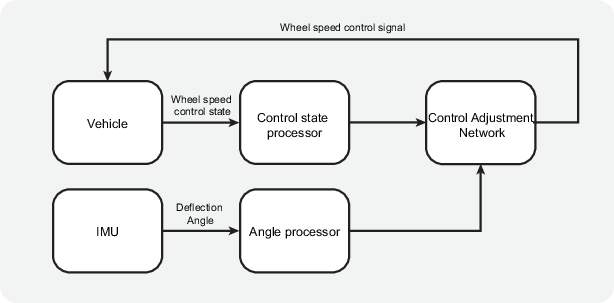}
\caption{The wheeled robots control flow.}
\label{fig:Wheel_speed_control}
\end{figure}

While the PID algorithm is commonly used for wheel speed control, it is not ideal for complex environments due to its cumbersome parameter tuning. Spiking neural networks (SNNs), combined with Inertial Measurement Units (IMUs), offer a low-power, adaptive solution for maintaining straight-line motion. Pulse Width Modulation (PWM) is employed to control the speed of the wheeled robots, with PWM signals scaled to fit the SNN model. Angle processing follows a similar rescaling method, mapping the angle to adjust robot direction. The overall control flow is shown in Fig. \ref{fig:Wheel_speed_control}.

SNNs use supervised learning algorithms like Prescribed Error Sensitivity (PES) \cite{voelker2015solution} or homeostatic PES (hPES) \cite{Bekolay2013Simultaneous}, combined with unsupervised methods like BCM, to tune control signals. These algorithms help the system adapt and maintain accurate motion control.

\subsection{Directional coding based on spiking neural network}

The continuous attractor neural network (CANN) is used to encode continuous variables like orientation and motion direction \cite{wu2008dynamics}. Inspired by biological evidence of head orientation neurons \cite{kim2017ring}, this model encodes the wheeled robot's orientation and direction of movement by forming a wave packet in the neural population. This approach allows the robot to respond effectively to the brain’s encoding of continuous variables.

\begin{figure}[ht]
\centering
\includegraphics[width=\linewidth]{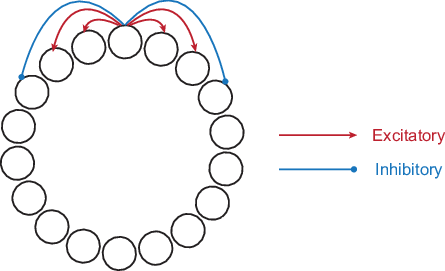}
\caption{The structure of attractor network, a network of nodes, often recurrently connected, whose time dynamics settle to a stable pattern.}
\label{fig:attractor}
\end{figure}

\section{Experiment}
This section presents experiments to evaluate the proposed cooperative shared control system for BMI tasks, focusing on object detection, robotic arm movement, and wheeled robot speed and orientation control using SNN.

\begin{figure}[ht]
\centering
\includegraphics[width=\linewidth]{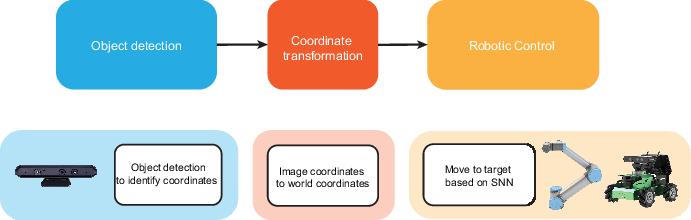}
\caption{The overall framework of the control system.}
\label{fig:robotic_framework}
\end{figure}

We first cover object detection for both robotic arms and wheeled robots, followed by robotic arm movement control using a UR5 model, and conclude with wheel speed and orientation control experiments. The overall control framework is shown in Fig. \ref{fig:robotic_framework}.

\subsection{Object detection}
We employed the YOLO algorithm for object detection tasks in both the robotic arm and wheeled robots. For the robotic arm, we used the Grape dataset, while for the wheeled robots, the Masked Face dataset \cite{ge2017detecting} was applied, containing images with and without face masks. Data augmentation techniques such as random cropping, flipping, contrast adjustment, and noise addition were used to enhance the dataset. The dataset was split 9:1 for training and testing.

\begin{figure}[ht]
\centering
\includegraphics[width=\linewidth]{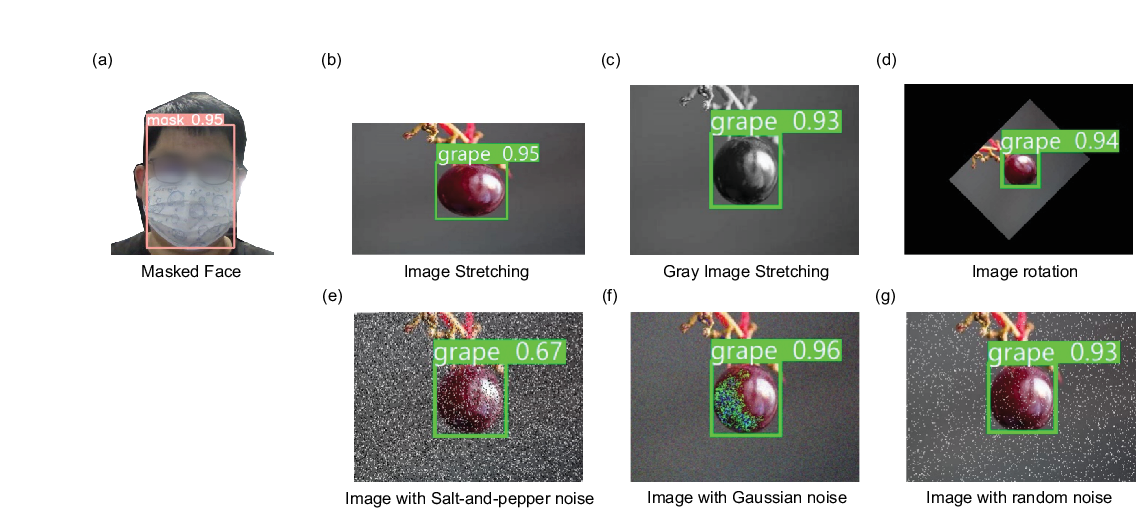}
\caption{The object detection based on YOLOv5s.}
\label{fig:YOLO_Detect}
\end{figure}

Robustness tests were conducted to evaluate object detection under varying conditions, including brightness, occlusion, rotation, stretching, and noise, as shown in Fig. \ref{fig:YOLO_Detect}. The model demonstrated effective recognition of both Masked Face and Grape objects with good robustness.

\subsection{Robotic arm movement control experiments}
\begin{figure}[ht]
\centering
\includegraphics[width=\linewidth]{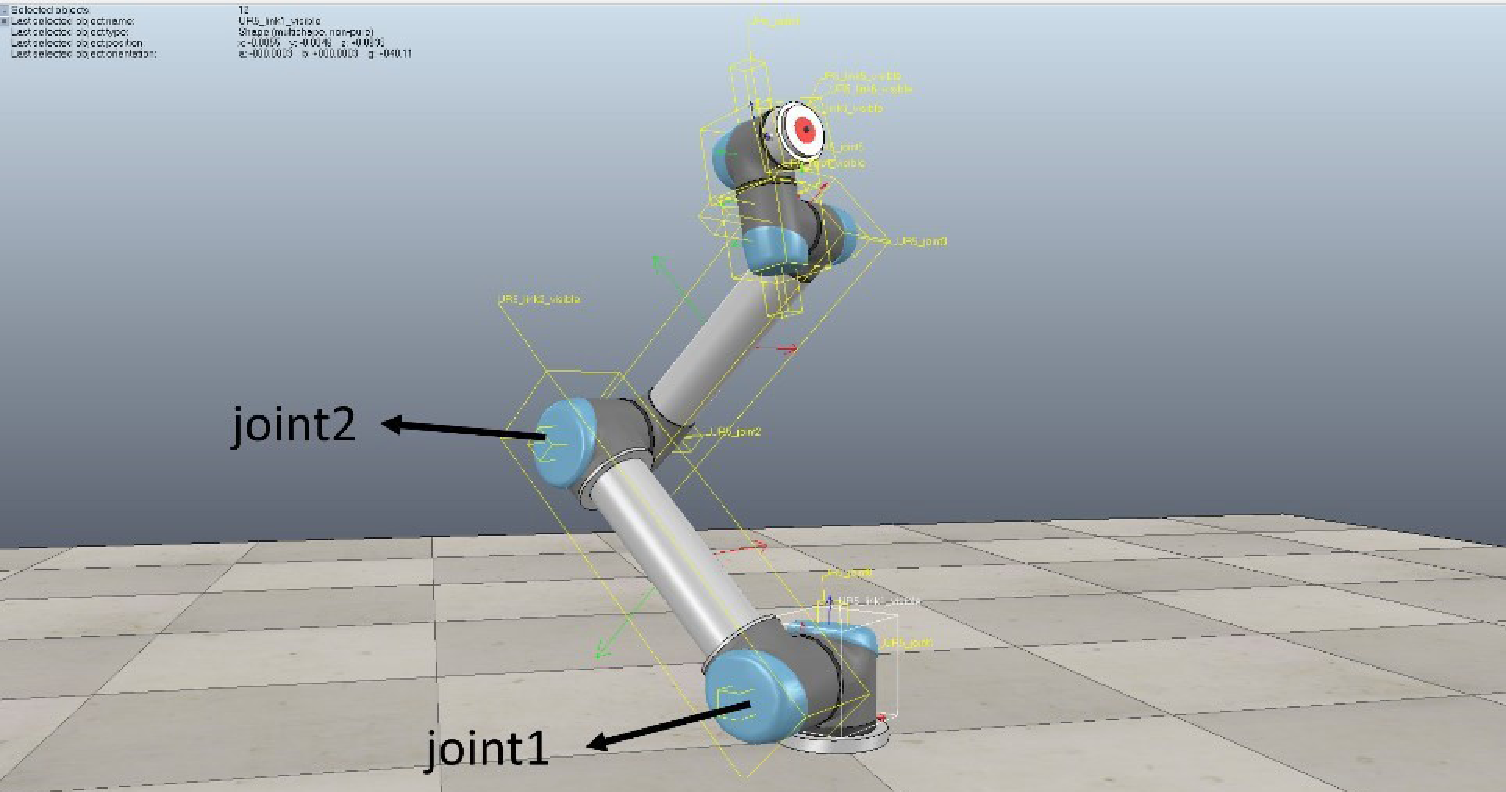}
\caption{The UR5 robotic arm runs on the simulation software CoppeliaSim.}
\label{fig:simulation_robotic_arm}
\end{figure}

Based on the work by Travis et al. \cite{dewolf2016spiking}, we implemented spiking neural network-based control for a UR5 robotic arm using CoppeliaSim for simulation and Nengo for neural network control \cite{bekolay2014nengo}, as shown in Fig. \ref{fig:simulation_robotic_arm}. The UR5 arm, consisting of six joints, was tested under realistic force conditions simulating unexpected disturbances, with joint1 and joint2 being the most critical.

We compared the performance of conventional PID control with the SNN-based control. The UR5 arm was tasked with moving from a starting position to four target positions at varying distances ($\Delta$X, $\Delta$Y, $\Delta$Z). For each direction, 20 simulations were conducted, comparing joint arc variations and the distance between starting and target positions. Results showed the differences in control performance between the two algorithms. 

\subsection{Wheeled robots speed control and direction orientation experiment}

For wheeled robot movement control using a brain-machine interface, we utilized the Robot Operating System (ROS) to integrate the wheel speed, direction control, and target detection modules. When the camera detects an unmasked object, the spiking neural network module directs the robot toward the object. ROS distributed communication connects the object detection and SNN modules.

The experiments were conducted with a Raspberry Pi-based two-wheeled robot using N20 motors and an Inertial Measurement Unit (IMU) to monitor heading angle. SNN-based wheel speed control, built using Nengo \cite{bekolay2014nengo}, applied homeostatic Prescribed Error Sensitivity (hPES) to adjust wheel speeds in real-time.

\begin{figure}[ht]
\centering
\includegraphics[width=\linewidth]{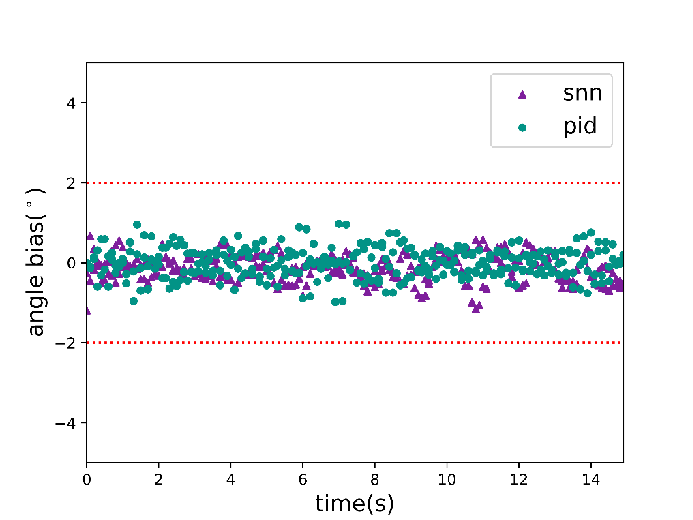}
\caption{Comparison of wheel speed regulation with different algorithms.}
\label{fig:hPES}
\end{figure}

Tests on uneven ground showed that the SNN effectively corrected directional deflection, with angular oscillation around 0 degrees, performing similarly to the classical PID algorithm, as shown in Fig. \ref{fig:hPES}.

\begin{figure}[ht]
\centering
\includegraphics[width=\linewidth]{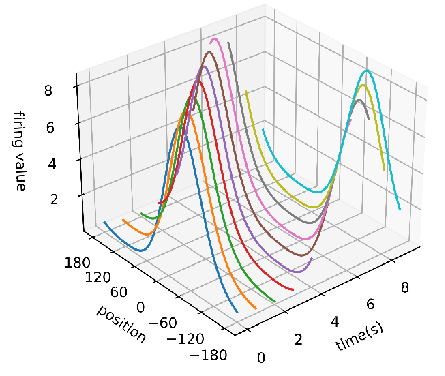}
\caption{The directional encoding based on the attractor network.} 
\label{fig:smooth_movement_attractor}
\end{figure}

For direction encoding, an attractor network based on Si Wu et al.'s design \cite{wu2008dynamics} was implemented with 360 neurons, each connected to 90 neighboring neurons, as shown in Fig. \ref{fig:attractor}. The network effectively encoded orientation information, as demonstrated in Fig. \ref{fig:smooth_movement_attractor}.

\section{Results}
In this case study, we aim for the robot to execute various functions accurately and robustly based on discrete instructions, performing more efficiently than traditional algorithms.

\subsection*{Robotic arm movement control}
\begin{figure}[ht]
\centering
\includegraphics[width=\linewidth]{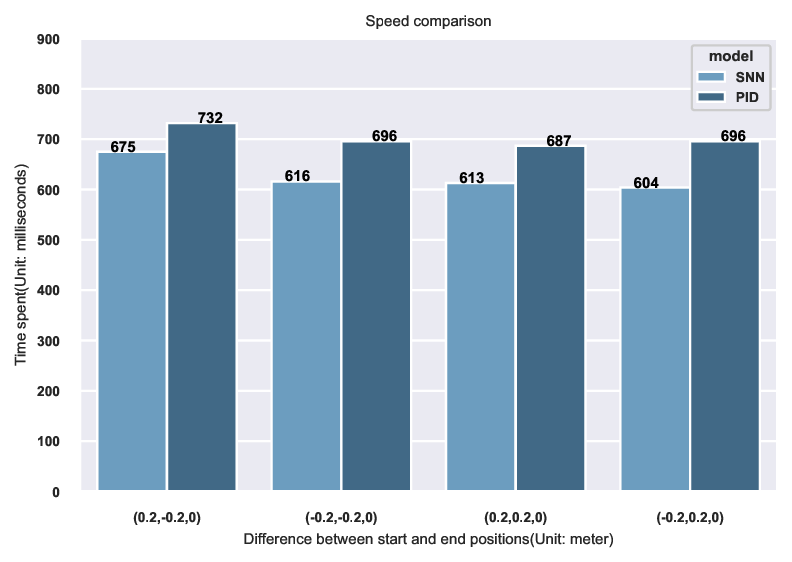}
\caption{Comparison of robotic arm movement speed with different algorithms.}
\label{fig:speed4}
\end{figure}

\begin{figure}[ht]
\centering
\includegraphics[width=\linewidth]{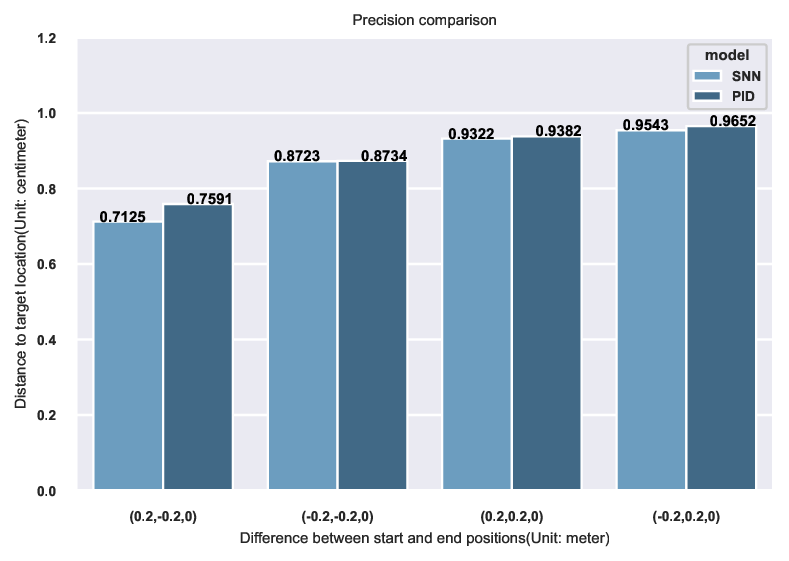}
\caption{Comparison of robotic arm movement precision with different algorithms.}
\label{fig:precision4}
\end{figure}

We tested the robotic arm's performance under accidental disturbances using both PID and spiking neural network control algorithms. A force between 1/4 and 1/16 of the arm's gravity was applied to simulate real-world conditions. The time taken to reach the target position and the distance to the target were measured (Fig. \ref{fig:speed4}, \ref{fig:precision4}).

The SNN algorithm enabled the robotic arm to approach the target faster than the PID algorithm (by more than 10\%), and while the final accuracy differences were minimal, the SNN provided faster stabilization under unexpected forces. This demonstrates that the SNN control algorithm better supports brain-machine interface tasks by improving response time under disturbance.

\subsection{Wheeled robots Object Tracking}

\begin{figure}[ht]
\centering
\includegraphics[width=\linewidth]{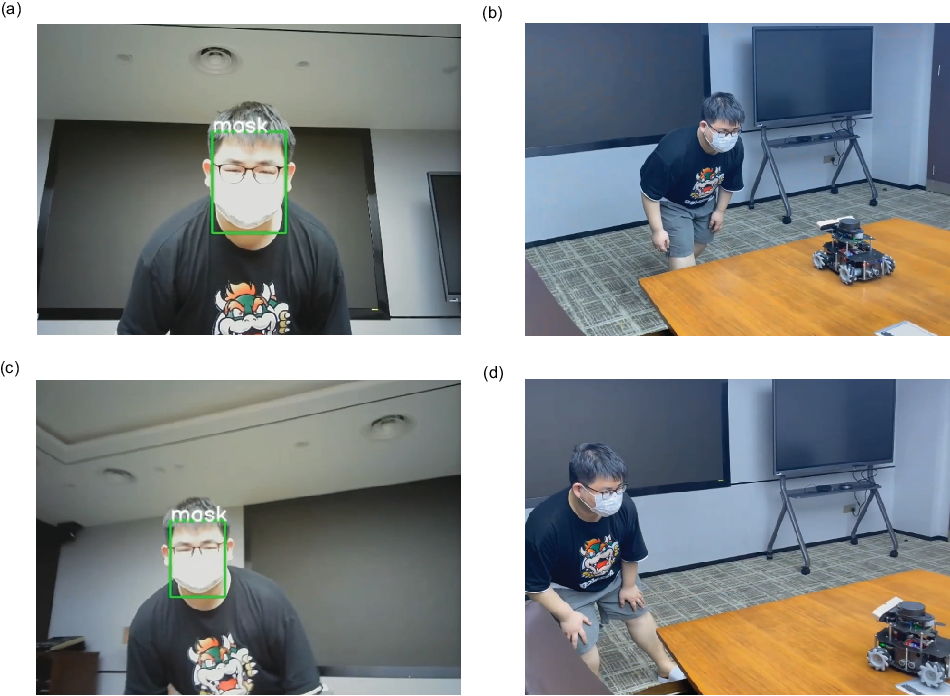}
\caption{The wheeled robot is tracked with Masked-Face as the target. (a), (c) are the images taken by the camera. (b), (d) are wheeled robot for target tracking.}
\label{fig:wheeled_robots_tract}
\end{figure}

\begin{figure}[ht]
\centering
\includegraphics[width=\linewidth]{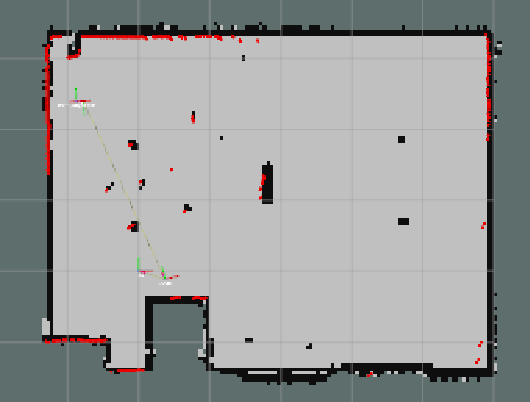}
\caption{The semantic map generation based on SLAM algorithms}
\label{fig:Map_generation}
\end{figure}

We combined the SNN control module with the object detection module on the wheeled robot for target tracking. The robot successfully followed a moving object in real-time (Fig. \ref{fig:wheeled_robots_tract}). Additionally, we used ROS's gmapping \cite{Quigley2009ROS} for SLAM during tracking, generating a detailed map of the environment (Fig. \ref{fig:Map_generation}).

\begin{figure}[ht]
\centering
\includegraphics[width=\linewidth]{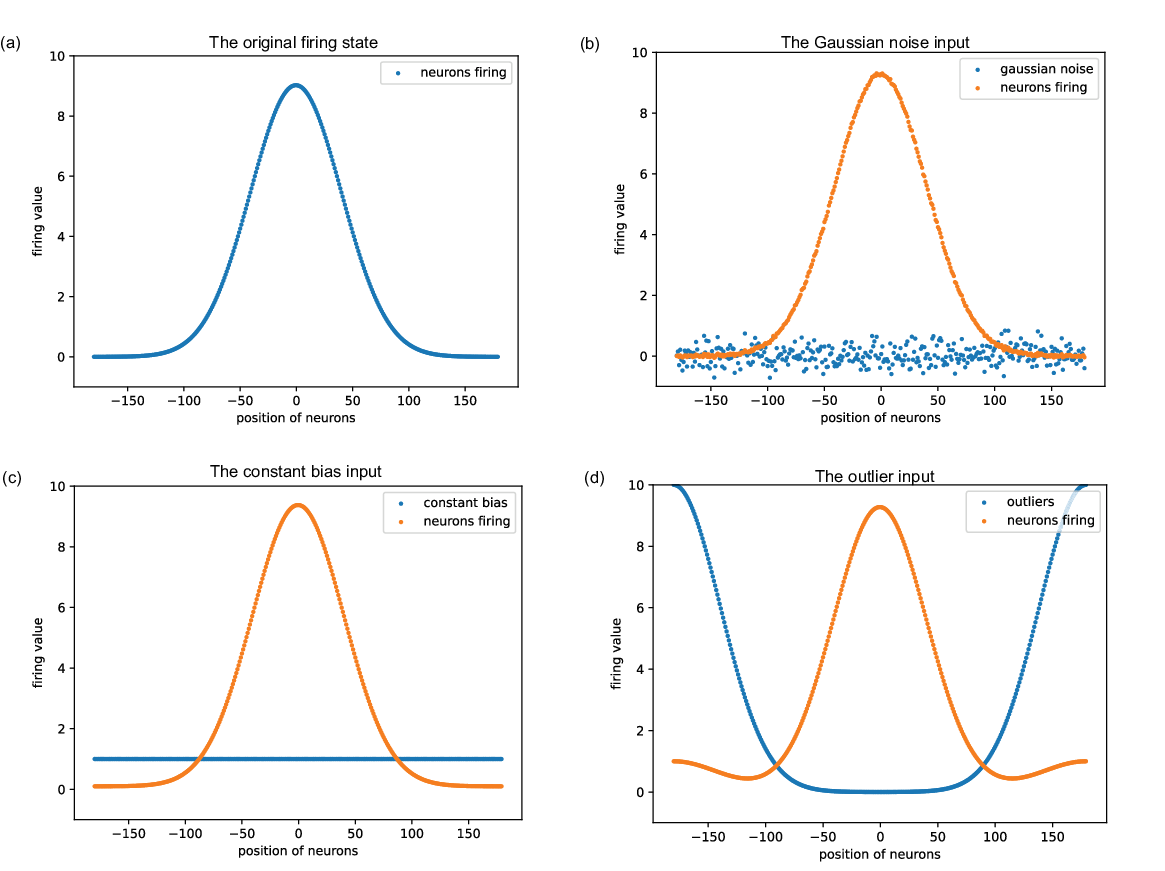}
\caption{The Gaussian noise, constant bias and outlier input to the original firing state of the continuous attractor as shown in (a), and observe the change of the release state of the attractor network, as shown in (b), (c) and (d).}
\label{fig:Robustness_test}
\end{figure}

We also tested the directional encoding of the wheeled robot using an attractor network under Gaussian noise. The results showed that the network maintained stability despite noise and anomalies, demonstrating its robustness (Fig. \ref{fig:Robustness_test}). The attractor network's stability under random inputs makes it suitable for navigation tasks, spatial cognition, and memory. The SNN's bio-inspired nature allows mobile robots to perform advanced tasks and directional movement effectively.

\section{Prospects and Challenges}

This study demonstrates the potential of brain-inspired intelligence in brain-machine interface (BMI) shared control systems for robotic arm and wheeled robot tasks. By integrating BMI signals with vision-guided control, complex tasks can be performed with adjustable autonomy. The system aims to maintain user control while assisting with more difficult tasks. Robots respond to user intent detected by BMI, optimizing task execution based on available signals.

Previous studies, such as Handelman et al. \cite{handelman2022shared}, achieved accurate control of modular prosthetic limbs for self-feeding tasks. However, manual adjustment of control parameters limits its scalability. Similarly, Dunlap et al. \cite{dunlap2019towards} demonstrated control of wheeled robots via motor cortex signals, but real-world challenges such as response time and fault tolerance remain.

Our BMI-based shared control system, tested in offline experiments, demonstrated improved performance in robotic arm movement and mobile robot direction control. The SNN-based control algorithm showed robustness to external forces and noise, offering better control compared to traditional methods.

Despite these advancements, challenges persist. Limited neural information and signal degradation restrict control performance, and SNNs have not yet shown clear superiority over artificial neural networks (ANNs) in certain tasks. Real-life applications, such as robotic grasping and navigation, involve additional complexities not seen in simulations. Visual feedback, for instance, may result in unstable grasping \cite{downey2016blending,kim2006continuous}, requiring further optimization.

The integration of spiking neural networks with biological motor control circuits, such as the cerebellum and brainstem, could enhance the bio-interpretability of robotic control. As hardware advances, such as neuromorphic chips and GPU acceleration, BMI systems are expected to become more portable and efficient. Future research should explore hybrid models combining the spatial-temporal advantages of SNNs with the data-driven capabilities of ANNs \cite{roy2019towards,lillicrap2020backpropagation}, potentially creating more generalized brain-inspired computing architectures for BMI applications.

This work serves as a proof of concept for applying brain-inspired intelligence to BMIs in shared control systems, striking a balance between user control and automation. As technology evolves, these systems may become invaluable tools for individuals with disabilities, significantly improving their quality of life.

\section{Conclusion}
This paper presents a collaborative shared control framework using brain-inspired intelligence and computer vision to control robotic arms and mobile robots. By distributing task execution between BMI signals and robotic control, efficiency and accuracy are enhanced. The results suggest that shared control systems hold great promise for practical BMI applications, potentially improving the daily lives of individuals with disabilities in the future.

\bibliography{sample}

\end{document}